\documentclass[10pt,a4paper]{article}
\usepackage[final]{lrec-coling2024}
\usepackage{latexsym}
\usepackage{multicol}
\usepackage{adjustbox}

\usepackage{footnotehyper}
\usepackage{forest}

\usepackage{microtype}

\usepackage{graphicx}

\makeatletter

\patchcmd{\NAT@test}{\else \NAT@nm}{\else \NAT@nmfmt{\NAT@nm}}{}{}

\DeclareRobustCommand\citepos
  {\begingroup
   \let\NAT@nmfmt\NAT@posfmt
   \NAT@swafalse\let\NAT@ctype\z@\NAT@partrue
   \@ifstar{\NAT@fulltrue\NAT@citetp}{\NAT@fullfalse\NAT@citetp}}

\let\NAT@orig@nmfmt\NAT@nmfmt
\def\NAT@posfmt#1{\NAT@orig@nmfmt{#1's}}

\makeatother

\usepackage{avm}
\usepackage{gb4e+}

\title{Spanish Resource Grammar version 2023}

\name{Olga Zamaraeva, Lorena S. Allegue, Carlos Gómez-Rodríguez} 

\address{Universidade da Coruña, CITIC, Departamento de Ciencias de la Computación \\ y Tecnologías de la Información\\
         Campus de Elviña s/n, 15071, A Coruña, Spain \\
         \{olga.zamaraeva, lorena.suarez, carlos.gomez\}@udc.es\\}

\abstract{
We present the latest version of the Spanish Resource Grammar (SRG), a grammar of Spanish implemented in the HPSG formalism. Such grammars encode a complex set of hypotheses about syntax making them a resource for empirical testing of linguistic theory. They also encode a strict notion of grammaticality which makes them a resource for natural language processing applications in computer-assisted language learning. This version of the SRG uses the recent version of the Freeling morphological analyzer and is released along with an automatically created, manually verified treebank of 2,291 sentences. We explain the treebanking process,  emphasizing how it is different from treebanking with manual annotation and how it contributes to empirically-driven development of syntactic theory. The treebanks' high level of consistency and detail makes them a resource for training high-quality semantic parsers and generally systems that benefit from precise and detailed semantics. Finally, we present the grammar's coverage and overgeneration on 100 sentences from a learner corpus, a new research line related to developing methodologies for robust empirical evaluation of hypotheses in second language acquisition.  \\ \newline \Keywords{grammars, treebanks, Spanish, HPSG, syntactic theory} }

\begin{document}
\maketitleabstract

\section{Introduction}
\label{sec:intro}

Among the various approaches to computational linguistics, formal grammars are a link between linguistic theory and natural language processing (NLP). By formal grammars we mean fully explicit linguistic formalisms encoding the general principles and operations involved in generating syntactic structure. Such formalisms are tied to fully-fledged theories of syntax and are developed by linguists independently of specific NLP needs or tasks. For example, Minimalism \citep{chomsky:minimalist}, Lexical Functional Grammar \citep[LFG;][]{kaplan1981lexical}, Head-driven Phrase Structure Grammar \citep[HPSG;][]{Pol:Sag:94} are linguistic theories of syntax. In contrast, the Penn-Treebank (PTB; \citealp{marcus1993building}, \citealplanguageresource{ptb-resource}) and the Universal Dependencies \citeplanguageresource[UD;][]{universaldep:2021} do not explicitly encode why something should be labeled in a certain way, and for that reason we consider them labeling conventions but not fully-fledged theories of syntax. It is hard to ensure that a labeling convention for complex structure is followed consistently. Formal grammars, on the other hand, not only encode complex linguistic hypotheses explicitly but, if implemented on the computer, map sentences to structures automatically and are fully consistent. Any error in them can be fixed systematically. Any previously labeled data can then be re-labeled automatically.  

Grammars take a long time to develop and the structures produced by them are harder to use than the annotation schemes developed specifically for NLP. For example, parsing can be much slower and the software stack generally needs to be more complex. However, grammars remain one of the few clear and long-term links between linguistics and NLP. In recent practice, formal implemented grammars have been used for computer assisted language learning (CALL) applications including grammar coaching \citep{flickinger2013toward, dacosta:2016:syntactic, dacosta:2020:automated}.\footnote{By grammar coaching, we mean detecting a grammar mistake and analyzing it linguistically to provide informed feedback rather than correcting the sentence that is considered wrong (grammar correction). Both grammar coaching and grammar correction are NLP tasks in the context of workshops such as Building Educational Applications \citep[BEA;][]{bea-2023-innovative}.} They have also been used to create high quality treebanks to train semantic parsers \citep{buys2017robust, chen2018accurate, lin2022neural}. In particular, \citet{lin2022neural} report a 35\% error reduction and 14\% absolute accuracy gain due to their use of the precise and consistent semantic representations generated by the English Resource Grammar \citep[ERG;][]{flickinger2000building, Flickinger:11}. In this paper, we present the latest version of a grammar which can be used to create such high quality training data for Spanish.

The Spanish Resource Grammar \citep[][]{marimon2010spanish, marimon2014automatic} is the second biggest implemented HPSG grammar  (see \S\ref{sec:delphin}). The latest version that we present uses a newer morphophonological analyzer through a completely reimplemented, easily editable Python interface. With this new interface, it is possible to use the grammar with the state-of-the-art HPSG parsers, tailoring it as necessary. We report the SRG's accuracy on a portion of the AnCora/TIBIDABO corpus \citelanguageresource{taule2008ancora, marimon2010tibidabo} for the first time and present an example of using a learner corpus to find areas for improvement in the grammar's encoded analyses.

We present work that is unusual in the sense that we are breathing new life into a valuable resource which remained dormant for at least 10 years. Unlike other software, grammars do not become obsolete in the sense that they encode robust linguistic theories. For that reason, we are convinced that the SRG should be reintegrated into the computational linguistics landscape, providing the community with a resource similar to the English Resource Grammar (ERG). Like other software, grammars do become obsolete since they depend on tools which may become outdated, and fixing such dependencies can be expensive. We present a year of work that went into enabling the SRG to work with a better parser and establishing its accuracy on 2K sentences\,---\,a time consuming process which has to be done once, before automatic tools can be leveraged to quickly compare new iterations. Building upon this foundation, the grammar can be expanded such that its coverage improves. 

The paper is organized as follows. In Section \ref{sec:bg}, we explain briefly the formalism behind the grammar implementation and what treebanking means in the context of grammar engineering. We also dedicate a section to crediting the original version of the grammar which took its author years to build. Section \ref{sec:improv-sum} describes what we did to bring the SRG up to date with the SOTA grammar engineering tools. Section \ref{sec:treebanks} is dedicated to evaluation. It gives an overview of a set of phenomena that are covered, as revealed by a specially constructed test suite (\S\ref{sec:mrs-test}); presents the results of the parsing with the grammar of 2,291 sentences from a Spanish news corpus (\S\ref{sec:tibidabo}) along with the discussion of the issues that the evaluation reveals; and of 100 sentences from a Spanish learner corpus (\S\ref{sec:learner}). Section \ref{sec:learner} includes an example of how the implemented grammar helps study syntactic hypotheses rigorously. The example shows a tension in the analyses that would likely remain overlooked if one did not implement them on the computer and did not run the grammar on a corpus containing ungrammatical examples.  

\section{Background and Methodology}
\label{sec:bg}

This section describes the formalism used in the Spanish Resource grammar (\S\ref{sec:delphin}-\S\ref{sec:hpsg}), related work, and the history of the Spanish Resource Grammar project (\S\ref{sec:srg-bg}). It explains how grammar-based treebanking is different from using labeling conventions for manual annotation (\S\ref{sec:treebanking}).

\subsection{Grammar engineering and DELPH-IN}
\label{sec:delphin}

Grammar engineering is a discipline for implementing syntactic theory on the computer. The ultimate research goal of grammar engineering is to refine syntactic theory in a general way and improve our systematic understanding of how human language works. A grammar implementation is a set of files. Parsers take such grammar implementations as input along with the sentences to parse the sentences according to the hypotheses encoded explicitly in the grammar files. As already mentioned, the theories underlying the formalisms used in grammar engineering encode the general principles hypothesized for syntactic structure by syntacticians. The associated level of complexity means they are less easy to use for NLP tasks but have bigger generalizability potential compared to labeling conventions.

There are several grammar engineering initiatives couched in various formalisms \citep{butt2002urdu, muller2015coregram, collins2016formalization}. DELPH-IN (DEep Linguistic Processing with Hpsg INitiative; \citealt{Copestake:02:CLE})\footnote{\url{https://delph-in.github.io/docs/home/Home/}} stands out as one with active international collaborations, an annual summit, and an emphasis on practical applications. The English Resource Grammar  \citep[ERG;][]{flickinger2000building, Flickinger:11}
 is the largest engineered grammar we are aware of (including outside of DELPH-IN). It empowered the creation of a large high-quality treebank originally published as \citealt{oepen2004lingo} with regular updates with each ERG release.\footnote{\url{http://svn.delph-in.net/erg/tags/2023}} Another unique initiative within DELPH-IN is the Grammar Matrix \citep{Ben:Fli:Oep:02, Ben:Dre:Fok:Pou:Sal:10, zamaraeva202220}, a system for automatic grammar creation based on typological description. The Grammar Matrix outputs grammar fragments which can then be developed further. DELPH-IN projects include grammars of Japanese, Chinese, Singaporean English, Hausa, German, Indonesian, Norwegian, Portuguese, Bulgarian, and more.\footnote{\url{https://delph-in.github.io/docs/grammars/GrammarCatalogue/}} A grammar of Spanish of a non-trivial size was only implemented in the DELPH-IN formalism, to our knowledge.\footnote{The CoreGram project \citep{muller2015coregram} included a starter Spanish grammar but, as far as we know, focused on other languages later on.}. 


\begin{figure*}[h!]
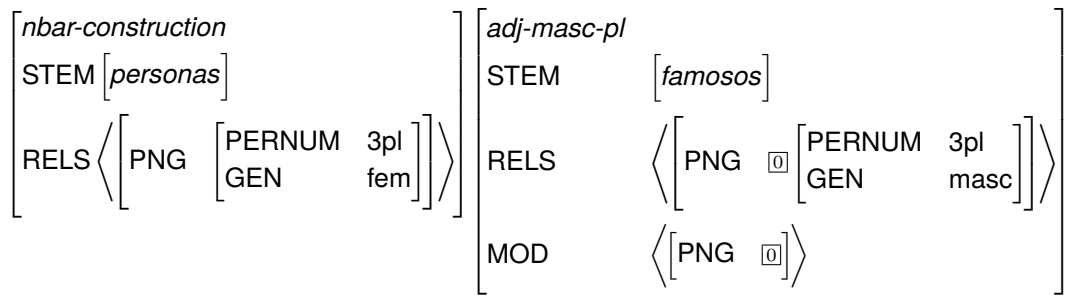

    \centering
\begin{avm}
\[\textit{nbar-construction} \\ 
STEM  \[ \emph{personas} \]  \\ 
RELS  \< \[ PNG & \[ PERNUM & 3pl \\ GEN & fem \] \]\> \]
\end{avm}
\begin{avm}
\[\textit{adj-masc-pl} \\
STEM & \[ \emph{famosos} \]  \\
RELS & \< \[ PNG & \@0 \[ PERNUM & 3pl \\ GEN & masc \] \]\> \\
MOD & \< \[ PNG & \@0 \] \> 
 \]
\end{avm}
    \caption{Two abbreviated feature structures produced by the SRG. Note the incompatible gender.}
    \label{fig:feature-structures}
\end{figure*}

\subsection{HPSG and MRS}
\label{sec:hpsg}

DELPH-IN grammar engineering uses the HPSG and the MRS formalisms.\footnote{For a more detailed overview of the relationship between the HPSG theory and computational linguistics, see \citealt{bender:emerson:handbook}.} Head-driven Phrase Structure Grammar \citep[HPSG;][]{Pol:Sag:94} is a constraint-based unification theory of syntax \citep{carpenter1992logic}. The formalism is fully explicit and serves as the foundation for multiple grammar engineering initiatives. HPSG sees syntactic structures as a hierarchy of phrasal and lexical types which can be instantiated as graphs containing feature-value pairs. The type hierarchy determines which values are compatible (can unify) and which are not. During unification-based parsing \citep[e.g.][]{carroll:1993, callmeier2000pet, crysmann2012towards, slayden2012array}, first, lexical analysis is performed and then a parse chart is built bottom-up, attempting to account for the entire input string with one feature structure that is compatible with the \emph{root conditions} (a set of constraints defining a full sentence). If something in a candidate feature structure cannot unify, the structure is discarded. Two simplified HPSG structures are presented in Figure \ref{fig:feature-structures}. They correspond to two words from an example from a learner corpus (\ref{ex:spa1}). 


\begin{exe}
\ex \gll *Mis abuelos son personas famosos.\\
my.{\sc 3pl} grandparent.{\sc masc.pl} be.{\sc 3pl.pres.ind} person.{\sc fem.3pl} famous.{\sc masc.pl}\\
Intended: `My grandparents are famous people.' [spa; \citealt{yamada2020cows}]
\label{ex:spa1}
\end{exe}

\begin{figure}
\centering
\begin{forest}
[...[\emph{png}[\emph{gender}[\emph{masc}][\emph{fem}][\emph{neut}]][\emph{pernum}[\emph{3sg}][...]]]]
\end{forest}
    \caption{A portion of the type hierarchy}
    \label{fig:hier}
\end{figure}
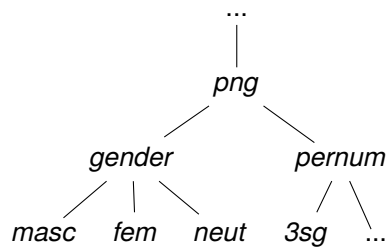

These structures (simplified for presentation) illustrate that two words have incompatible agreement values and so could not be used to form a noun phrase. The structure for the adjective (right) specifies an identity between the person, number, and gender ({\sc PNG}) features of the adjective and the word that it modifies (the identity is represented as the numbered tag \avmbox{0}). The values such as \emph{3pl} and \emph{fem} come from the type hierarchy (Figure \ref{fig:hier}) while the orthographies come from the lexicon, in this case paired with an external morphological analyzer.
Since \emph{fem} and \emph{masc} do not unify,\footnote{The fact that they do not unify is determined by the type hierarchy; here we omit the detailed explanation of the mechanics of unification which can be found in e.g.\ \citealt{Copestake:02} and \citealt{Copestake:02:CLE}.} the structure on the left could not possibly be on the {\sc mod} list of the structure on the right. 

While an HPSG structure can be visualized as a tree (as shown later in Figure \ref{fig:tree}), in reality it is a more complex graph which includes full information on all the constraints (the tree only includes node labels which are not meaningful on their own and serve only for exposition). A graph for a full sentence can be visualized as an attribute-value matrix like the ones in Figure \ref{fig:feature-structures} but with the full set of constraints arising from the complete grammar. The \emph{type} of such a structure is a phrasal type rather than a lexical type (phrasal and lexical types all belong to the same type hierarchy which is partially shown in Figure \ref{fig:hier}).

While Figure \ref{fig:feature-structures} shows a simple example of gender agreement, HPSG can model syntactic complexity in full detail. This is particularly useful when semantic nuances accessible through the syntax-semantics interface matter. HPSG has been used to solve issues related to negation scope \citep{packard2014simple, zamaraeva2018improving} and compositionality generally \citep{lin2022neural}.

\begin{figure*}
    \includegraphics[width=\textwidth]{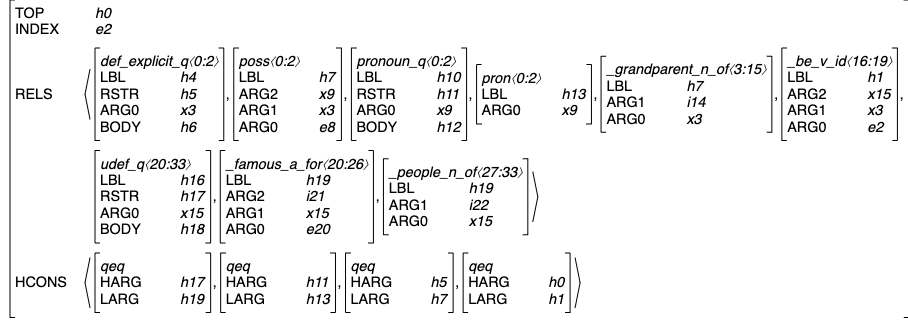}
    \caption{MRS for the sentence \emph{My grandparents are famous people}. The main event is labeled \emph{e2}; its dependencies are \emph{x3, x15}. RSTR is used to track the scope of quantifiers and modification.}
    \label{fig:mrs}
\end{figure*}

\begin{figure*}
        \includegraphics[width=\textwidth]{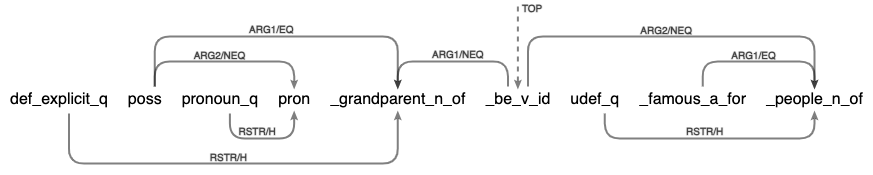}
    \caption{Dependency MRS for the sentence \emph{My grandparents are famous people}.}
    \label{fig:dmrs}
\end{figure*}

Semantics in DELPH-IN is modeled via the Minimal Recursion Semantics formalism \citep[MRS;][]{copestake2005minimal}. Any HPSG structure includes semantic constraints. Such constraints are partially shown in Figure \ref{fig:feature-structures} as {\sc rels} but a non-simplified structure produced by a DELPH-IN grammar includes a full MRS. An MRS is a bag of predications which include information about various semantic properties of the structure, including quantifier, negation, and modification scope; tense and aspect of events; person, number, and gender of entities, and information structure. The MRS for the intended meaning of sentence (\ref{ex:spa1}) is given in Figure \ref{fig:mrs}. An MRS can be automatically converted to a dependency structure (Figure \ref{fig:dmrs}).\footnote{Both figures were generated by the DELPH-IN online demo: \url{http://delph-in.github.io/delphin-viz/demo}.}

\subsection{The Spanish Resource Grammar}
\label{sec:srg-bg}
The Spanish Resource Grammar (SRG) \citep{marimon2010spanish, marimon2014automatic} is the second biggest DELPH-IN grammar. It has 226 phrase structure types, 504 lexical rule types, 543 lexical types, and a lexicon of 54,510 lemmas.\footnote{We would like to stress that the SRG was originally developed not by us and we have the big privilege to build on this substantial prior research that took years of effort by \citeauthor{marimon2010spanish} and her colleagues.} The morphophonological analysis is done externally by Freeling \citelanguageresource{padro2012freeling, carreras2004freeling}.\footnote{\url{https://nlp.lsi.upc.edu/freeling/}} An input sentence is first run through Freeling which outputs one or more possible lemma-tag pairs for each word. The Freeling output is passed to the parser. The parser is a separate tool which takes a DELPH-IN grammar as input along with the input sentence. The parser maps the lemmas to the lexical entries in the lexicon and the tags to the lexical rules. The lexical rules are designed to ensure that the word is analyzed as an HPSG feature structure  with the appropriate feature values such as specific values for gender, number, etc. When the SRG was first developed, the parser used was the PET parser \citep{callmeier2000pet}. It has since stopped being supported but the grammar was left with a dependency in the form of the Freeling-parser interface.

SRG's accuracy\footnote{\textbf{Accuracy:} how many grammatical sentences get assigned the desired semantic structure. This is one of the main metrics for grammar evaluation.} with respect to any corpora was never published (as far as we can tell). The coverage\footnote{\textbf{Coverage:} how many grammatical sentences are assigned some (any) constituency structure by the grammar, including not only correct but also spurious structures (leading to wrong or broken semantics). This metric is relevant for grammar development but less so for ultimate evaluation.} for 17K sentences from AnCora \citelanguageresource{taule2008ancora} with the PET parser was reported in \citealt{marimon2010spanish} and \citealt{marimon2010tibidabo} but these coverage figures include undesirable structures and were obtained with a slower parser that would time out often. We have obtained higher coverage (92\% vs.\ 74\%) thanks to the better parser speed and have manually verified the accuracy for sentences up to and including length 10.\footnote{In this process, we had access to treebanking decisions made by \citeauthor{marimon2010tibidabo} but we could not directly incorporate them due to the Freeling versioning incompatibility.} 

\subsection{DELPH-IN Treebanking}
\label{sec:treebanking}

In the context of DELPH-IN grammars, treebanking is in a sense the opposite to the treebanking in the settings such as Universal Dependencies (UD). Treebanks like UD are created manually with the goal to then train statistical tools on them. Conversely, DELPH-IN treebanks are created automatically by the manually built grammar. While it is clear that creating treebanks automatically is generally preferable due to higher consistency and scalability, there are two caveats: (1) language is highly ambiguous; (2) the grammar is not perfect. This means the grammar may generate many structures for each sentence, some of which may be semantically implausible or plain wrong. Therefore, the treebank generated automatically has to be gone through manually at least once, to pick the correct/desired tree and record it as the ``gold'' result for the particular sentence, or to record any ``bugs'' in the grammar that need to be fixed. The verification is done with respect to the semantic structure (MRS; Figure \ref{fig:mrs}); in the context of DELPH-IN grammars, the specifics of the constituency structure are secondary as long as they lead to the correct semantics.\footnote{This is characteristic of a syntax theory in development. We assume that we do not yet have a full understanding of what the complete set of correct syntactic analyses is, so we assess them via the correctness of the semantics that they yield.} 

Manual treebanking in DELPH-IN is the necessary step to train parse selection models required in most realistic applications. Human language is ambiguous, and the desired semantic structure is often determined only by pragmatics. That is outside of the scope of a syntactic theory, and an HPSG grammar will dutifully produce all the structures that it considers \emph{syntactically} possible, and statistical tools are required to choose the \emph{pragmatically} best one.\footnote{It is worth noting that pragmatically less plausible structures may still be meaningful, and there may be scenarios where producing them is a desideratum. A purely statistical system would be poor at such a task.} 

The recorded gold results from a treebank can be automatically compared to a parse forest that represents new results in the next iterations of the grammar development. In other words, if a bug is fixed in the grammar or a new analysis is added, the impacts of the change can be assessed by the number and types of differences resulting from running the grammar on the same set of sentences and automatically comparing the output with the recorded gold. 

The process of picking the gold tree from the full parse forest is time consuming but it is still faster and more consistent than creating a treebank manually from scratch.\footnote{Of course, the reverse is true about the grammar: it takes a lot of time to build, compared to a statistically trained resource.} Also, the bigger the treebanks, the better the parse ranking model, and once we are confident enough of the parse ranking and the grammar quality, we can parse new data and use it without verification. 

In this section, we described the grammar engineering methodology including treebanking, a way of annotating corpora for syntactic structure consistently and semi-automatically. The next section summarizes the improvements we introduced to the SRG, in particular to be able to grow the treebanks more efficiently.

\section{Summary of improvements}
\label{sec:improv-sum}
In this iteration of the SRG development, we achieved four main objectives: (1) have it work with the parsers ACE \citep[][with regular releases since 2012]{crysmann2012towards},\footnote{\url{http://sweaglesw.org/linguistics/ace/}} and the recent open-source version of the parser LKB \citep[][with regular releases since the publication date]{carroll:1993};\footnote{\url{https://delph-in.github.io/docs/tools/LkbFos/}} (2) have it use a recent version of the Freeling morphophonological analyzer, v4.1;\footnote{\url{https://nlp.lsi.upc.edu/freeling/index.php/node/1}} (3) establish the current coverage and accuracy of the grammar on (a portion of) the AnCora corpus; (4) use the grammar on a learner corpus, as a step towards using it in CALL applications and to better understand the current grammar's overgeneration.\footnote{\textbf{Overgeneration:} outputting wrong structures for grammatical sentences or any structure for an ungrammatical sentence.} Adding new analyses to the grammar for it to support more phenomena is future work; first we needed to establish where it is now.

To reach these objectives, we have (1) revised the portion of the grammar responsible for the inflectional lexical rules to match Freeling v4.1 morphophonological analyzer tagset; (2) implemented a new, modular Python interface between Freeling, the grammar, and the ACE and LKB parsers;\footnote{We are indebted to John Carroll for implementing several required modifications in the open-source version of the LKB parser.} (3) re-parsed the data up to length 20 with the ACE parser; (4) semi-manually verified the accuracy of the grammar on the sentences up to and including length 10; (5) explored the current grammar coverage and accuracy and documented them in the form of GitHub issues;{{\interfootnotelinepenalty10000 \footnote{\url{https://github.com/delph-in/srg/issues}}}
(6) prepared a new dataset based on an existing learner corpus to explore the grammar's overgeneration. In summary, we present a version of the grammar which is ready to use with SOTA DELPH-IN parsers and 
whose coverage and limitations are clearer.
The version of the SRG corresponding to this paper can be found on GitHub under Releases v0.3.3.\footnote{\url{https://github.com/delph-in/srg/releases/tag/v0.3.3}}. 
The release includes the treebanks.




\section{Grammar evaluation}
\label{sec:treebanks}
In this section, we present an assessment of the SRG that we performed for the first time thanks to the improvements summarized in \S\ref{sec:improv-sum}. Generally speaking, grammars can be deficient in two ways: they can lack accuracy (not provide a correct structure for a sentence) or they can overgenerate (provide a wrong structure for a sentence). Accuracy and overgeneration are therefore the two principal metrics we use to evaluate grammars. We start from targeted evaluation of the accuracy on a set of constructed illustrative examples of linguistic phenomena (\S\ref{sec:mrs-test}) and then present a larger-scale assessment of the accuracy on the AnCora/TIBIDABO corpus (\S\ref{sec:tibidabo}). Measuring overgeneration can be harder, since corpora of ungrammatical examples are not common. Overgeneration can be estimated indirectly through  noticing excessive ambiguity; if the grammar yields thousands of structures for a sentence, it is usually a sign of overgeneration, because even though natural languages are highly ambiguous, it is usually not the case that one sentence has thousands of possible meanings. We give the SRG ambiguity figure and comment on it at the end of \S\ref{sec:tibidabo}. In \S\ref{sec:learner}, we suggest using a learner corpus for studying overgeneration, although what we present in this paper is merely a starting point. 

\subsection{The MRS test suite}
\label{sec:mrs-test}

The MRS test suite is a collection of sentences illustrating semantic phenomena that are accessible through syntax.\footnote{\url{https://github.com/delph-in/docs/wiki/MatrixMrsTestSuite}} It is a way to assess a grammar's quality with respect to a range of linguistic phenomena by examining the adequacy of the MRS representations of the sentences illustrating the phenomena that the grammar yields. Across languages, the MRS structures for the listed sentences will in some cases be similar and in others they will not be, depending on how differently the languages in question express certain meanings. The test suite was first compiled for English in the context of the ERG development, and the English suite consists of 107 sentences.\footnote{\url{https://github.com/delph-in/docs/wiki/MatrixMrsTestSuiteEn}} The phenomena include different kinds of dependencies, scope of negation, scope of modification, implicit arguments (ellipsis), interrogatives, imperatives, and so on. The expectation is that we can compile a similar test suite for any language, as we expect to find such phenomena in most languages of the world. We also expect some differences because languages vary in the degree to which certain semantic phenomena are exposed through syntax. The semantically analogous sentences in the MRS test suites for different languages should correspond to each other by the number ID. The items which have no correspondence should have unique IDs. 

We had the MRS test suite for Spanish compiled for the original release \citep{marimon2010spanish}. We have edited the test suite to better reflect the facts of the Spanish language related to e.g.\  flexible word order interacting with focus. On the other hand, we corrected some mistakes where a Spanish sentence was identified as an equivalent to an English one where in reality the sentence had different semantics and thus should have been assigned a different ID. After adding some examples which seemed missing and removing some examples which seemed redundant,\footnote{Relevant discussion: \url{https://delphinqa.ling.washington.edu/t/abrams-wiped-the-table-clean-in-spanish/881}} the updated test suite consists of 106 sentences.\footnote{\url{https://github.com/delph-in/srg/blob/main/tsdb/txt-id/mrs/mrs-updated.txt}}

\begin{table*}[t!]
    \centering
    \begin{tabular}{llllc}
        sentence length & number of sentences & coverage & accuracy & times hit RAM limit \\
        1 & 65&1.0&1.0&0\\
        2 &177&0.94&0.94&0\\
        3 &181&0.91&0.89&0\\
        4 &219&0.91&0.86&0\\
        5 &229&0.92&0.87&0\\
        6 &211&0.91&0.83&0\\
        7 &246&0.91&0.76&0\\
        8 &278&0.93&0.82&0\\
        9 &326&0.92&0.78&5\\
        10 &359&0.91&0.76&3\\
        \hline
        all & 2291 & 0.92& 0.82 & 8 \\
    \end{tabular}
    \caption{SRG accuracy on the first 10 portions of the TIBIDABO treebank}
    \label{tab:tibidabo}
\end{table*}

The current SRG accuracy on the MRS test suite is 81\%. Examining the items for which the grammar did not yield a correct analysis has allowed us to document some areas where the grammar should be improved. Based on the results of running the grammar on the MRS test suite, we opened 11 new issues in the SRG GitHub repository including the ones related to: Missing analysis of imperfective and perfective aspect distinction in some cases; missing possessive relations in some cases; missing interrogative semantics in many cases (underspecification between a question and a proposition, which is expected in Spanish yes-no questions but not in e.g.\ \emph{wh}-questions); broken dependencies in some complex clauses including relative clauses and subordinate clauses, again, in some cases; insufficient implementation of the semantics associated with object clitics and the clitic \emph{se} (a structure similar to the correct structure is yielded by the grammar but the dependency between the subject and the clitic is broken). All of these issues are major but it is expected that the grammar does not yet handle all of them perfectly because it is still a relatively young grammar in terms of the time that went into its development so far. The point of the MRS test suite is to provide a good estimate of where the grammar is now and where to go next.

\subsection{TIBIDABO}
\label{sec:tibidabo}
The TIBIDABO treebank \citelanguageresource{marimon2010tibidabo} is a version of the AnCora corpus \citelanguageresource{taule2008ancora}  sorted by sentence length (to see the effects of sentence length on the HPSG parsing speed; see \S\ref{sec:parser-limitations}) and annotated for HPSG structure. \citetlanguageresource{marimon2010tibidabo} reports coverage (but not accuracy) on the sentences up to length 40. We were able to recover 5,894 annotated sentences representing sentence length 1-19, representing 33\% of the AnCora corpus. 
The rest of the TIBIDABO treebank appears to have been lost. We intend to rebuild it. 

For the 5,894 sentences we have recovered, we had parse forests which were partially verified for the gold standard. But since Freeling 4.0 updates resulted in some incompatibility with the previous version, we had to look at each and every tree again.\footnote{This does raise questions about the long-term desirability of the Freeling dependency; it may be possible to instead model the morphology directly in the grammar.} For the latest release, we have managed to examine the parse forests and verify the presence of the correct tree for sentences up to and including length 10 (2,291 sentences in total). Together with the Freeling interface overhaul, the verified portion of the treebank constitutes the main contribution of this paper.

Table \ref{tab:tibidabo} shows the results we have so far on TIBIDABO. The coverage is stable at 92\%\, although we expect it to go down as sentences become longer. The accuracy already goes down noticeably as the length goes up. Both coverage and accuracy suffer due to two main reasons: (1) parser limitations on long sentences, which could be overcome externally to the grammar (\S\ref{sec:parser-limitations}); and (2) genuine lack of the correct analyses encoded in the grammar (\S\ref{sec:coverage-issues}). 

\subsubsection{Parser limitations on longer sentences}
\label{sec:parser-limitations}

HPSG parsing is relatively slow. The goal of the parsing is exhaustive search in a large space of possible complex structures \citep{carroll:1993, crysmann2012towards}. With grammars which admit high ambiguity (see \S\ref{sec:ambig}), the size of the parse chart can quickly become prohibitive as 
sentence length
grows. This issue has been explored with respect to the ERG \citep{dridan2008enhancing, dridan2009using, dridan2013ubertagging, zamaraeva2023revisiting}, and similar solutions can be applied to the SRG. Once the treebanks become big enough, a statistical model can be trained to filter out unlikely edges from the 
chart. Currently, to parse a sentence of length 10 takes 1 second/sentence on average.

\subsubsection{Summary of genuine coverage issues}
\label{sec:coverage-issues}
Apart from the issues documented in relation with the MRS test suite, we have documented two groups of problems: (1) issues related to (possibly wrong) Freeling tags; (2) issues related to multiword expressions\,---\,not an easily solved problem because there is no universal treatment of MWE that would not involve trade-offs \citep{contreras2022models, sag2002multiword}. In addition, there are linguistic issues which do not immediately form a group and which call for individual investigation and possible reanalysis of portions of the lexicon and the type hierarchy.\footnote{The documented issues can be found here: \url{https://github.com/delph-in/srg/issues}.} 

Any change in the grammar may have wide effects on 
its
behavior with respect to data. For that reason, extending coverage requires rigorous testing. We present one example of lack of coverage and its preliminary investigation. The work for increasing the coverage is ongoing and new figures will be reported in future versions of the grammar.

\begin{exe}
    \ex \gll Mis amigos pueden venir si quieren.\\
    my.{\sc pl} friends can.{\sc 3pl.pres} come.{\sc inf} if want.{\sc 3pl.pres}\\
    `My friends can come if they want.' [spa]
    \label{ex:quieren}
\end{exe}

Example (\ref{ex:quieren}) and similar examples are not parsed by the grammar. What we discovered is, according to the grammar, the verb \emph{querer} (`to want') is assigned to a type which does not allow the kind of long-distance dependency that is required to form the sentence. In the sentence, the subject is shared between `come' and `want' and is not overtly present in the clause where `want' is the predicate. The issue is related to the overall complex analysis of long-distance dependencies in the grammar which was not fully finished (as far as we can tell). Developing the analysis will automatically improve not only the coverage with respect to sentence (\ref{ex:quieren}) but with respect to many more examples containing this kind of long-distance dependencies.

\subsubsection{Studying excessive ambiguity}
\label{sec:ambig}
The current version of the SRG has high ambiguity (482 structures per sentence, on average on the portions of TIBIDABO length 1-10). While natural languages including Spanish are highly ambiguous, having millions of structures per sentence (which is the case for some sentences) is clearly overgeneration. Such extreme figures are explained combinatorically; the longer the sentence, the more possibilities for different interpretations for each word and then each subconstituent containing each of those possibilities for each word. In some cases, this is inevitable and has to be sustained. In others, it may turn out that an additional constraint will preclude a number of chart edges from being hypothesized by the parser without any loss in the accuracy. The investigation for reducing ambiguity is ongoing work on which we do not report here.

\subsection{COWSLH2}
\label{sec:learner}
COWSLH2 is a  corpus of written Spanish learner language developed at UC Davis \citep{yamada2020cows}. The corpus contains over 100K sentences in the form of essays written by college students. Some sentences are annotated for grammatical errors. For the purposes of this paper, we semi-randomly selected 100 sentences of length up to 8. Of them, 36 are considered ``ungrammatical'' in the sense that they have some learner usage not characteristic of proficient Spanish speakers.\footnote{The third author, whose first language is Spanish, verified the grammaticality of the sentences.} The remaining 64 are grammatical sentences. We ran the SRG on the sentences. Ideally, we would like the SRG to parse only the 64 grammatical ones. As for the ungrammatical ones, 
ideally, we would expect it to reject them (not assign any structure). Of course, we know the SRG is not perfect, so the purpose of this exercise is to see where the room for improvement is.

\begin{table}[h!]
    \centering
    \begin{tabular}{lll}
        \hline
        coverage & accuracy & overgeneration \\ \hline
         100\% & 87\% & 61\% \\ \hline
    \end{tabular}
    \caption{SRG accuracy and overgeneration on 100 learner sentences}
    \label{tab:learner}
\end{table}

\begin{figure*}[h!]
    \centering
    \includegraphics[width=0.45\textwidth]{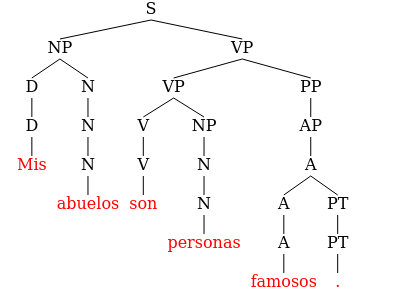}
    \includegraphics[width=0.45\textwidth]{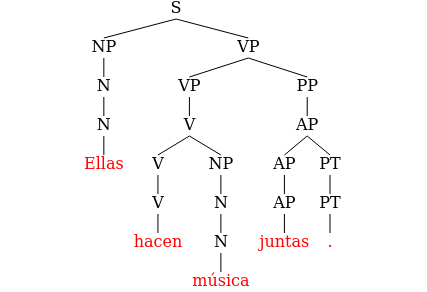}
    \caption{Left: A spurious tree for the Spanish sentence (\ref{ex:spa1-rep}). Right: A correct analysis for sentence (\ref{ex:spa2}). (Unary chains in HPSG may represent lexical rules as well as rules for e.g.\ argument drop.)}
    \label{fig:tree}
\end{figure*}

\subsubsection{Assessing overgeneration with a learner corpus}
\label{sec:example}

Table \ref{tab:learner} shows the results of running the SRG on the 100 short sentences from the learner corpus. The coverage is 100\% meaning all of the 64 grammatical sentences were assigned some HPSG structure. However, that does not mean the corresponding semantics is the desired one; the accuracy is 87\%. The large overgeneration figure (61\%) means that the grammar currently generates some structure for many ungrammatical sentences. 

The SRG's large overgeneration on the portion of COWSLH2 (61\%) is not unexpected; controlling for overgeneration requires regularly testing the grammar with ungrammatical sentences, which is done routinely in e.g.\ the Grammar Matrix project \citep{Ben:Dre:Fok:Pou:Sal:10} but, since larger grammars typically prioritize coverage over large corpora, overgeneration can grow. The bigger point here is that our ideas about how the grammar works are typically far from perfect and must be tested empirically and computationally. Following \citet{bierwisch1963grammatik}, \citet{butt1999grammar}, \citet{Bender:08}, \citet{fokkens2014enhancing}, \citet{muller2015coregram}, \citet{zamaraeva202220}, \emph{inter alia}, we emphasize that overgeneration 
and
other problems with the grammar are easy to overlook if one (1) does not implement the grammar and only considers sets of syntactic analyses in isolation and on paper; (2) only tests the grammar on cherry-picked examples. Running the grammar on a learner corpus is one method of assessing overgeneration.

Consider one example of how the learner corpus helps us quickly find problems in the grammar. 
\begin{exe}
\ex \gll *Mis abuelos son personas famosos.\\
my.{\sc 3pl} grandparent.{\sc masc.pl} be.{\sc 3pl.pres.ind} person.{\sc fem.3pl} famous.{\sc masc.pl}\\
Intended: `My grandparents are famous people.' [spa; \citealt{yamada2020cows}]
\label{ex:spa1-rep}
\end{exe}
The ungrammatical sentence (\ref{ex:spa1}) repeated here as (\ref{ex:spa1-rep}) is actually parsed by the SRG. Examining the assigned structure, we see that the adjective \emph{famosos} is attached high in the VP subtree, modifying the verb phrase rather than the noun (Figure \ref{fig:tree}).  The semantics of such a structure appears nonsensical.\footnote{The semantics is something like ``My grandparents are people while being famous."} However, disallowing adjectives from attaching high generally of course is not the solution; looking at where else this structure occurs in the corpus, we find ``healthy'' examples like (\ref{ex:spa2}), where the structure is sensible and necessary. 

\begin{exe}
\ex \gll Ellas hacen música juntas.\\
they.{\sc 3pl.fem} do.{\sc 3pl.pres.ind} music.{\sc fem.3sg} united.{\sc fem.pl}\\
`They play music together.' [spa; \citealt{yamada2020cows}]
\label{ex:spa2}
\end{exe}

What we find, then, is that in the SRG, the analysis of adjectives serving as verb modifiers applies too freely. But this is not the question of a simple reassignment of \emph{famoso} to a different lexical type. The adjective \emph{famoso} can be predicative (\ref{ex:spa3}), even if it cannot modify a head-complement construction such as \emph{son personas}.
\begin{exe}
\ex \gll Mis abuelos son famosos.\\
my.{\sc 3pl} grandparent.{\sc masc.pl} be.{\sc 3pl.pres.ind} famous.{\sc masc.pl}\\
`My grandparents are famous.' [spa]
\label{ex:spa3}
\end{exe}

In Spanish, there are two verbs \emph{to be}: \emph{ser} and \emph{estar}, and they are not interchangeable and convey different senses of being/state. A plausible hypothesis is that modified VP structures like in Figure \ref{fig:tree} are possible only with adjectives that occur with the verb \emph{estar} (e.g.\ \emph{junto}) and not with the ones that occur with \emph{ser} (e.g.\ \emph{famoso}). But it is not clear whether this distinction is ultimately not pragmatic.\footnote{In principle, while the meaning ``The grandparents are people while being famous'' is bizarre and does not sound like something anyone would say, perhaps there is a semantic universe in which it makes sense.} The question then is, how/whether to implement this distinction in the grammar and what effect will the changes have on the rest of the grammar, as evaluated not only with respect to (\ref{ex:spa1-rep})-(\ref{ex:spa3}) but to the entire corpus treebanked so far.  


\section{Conclusion and future work}
\label{sec:conclusion}
We presented the latest version of the Spanish Resource Grammar (SRG) and its accuracy over a portion of the TIBIDABO treebank. The grammar can be used in linguistic research and in computer-assisted language learning (CALL) applications. The treebank, as it grows in the future, can be used for training high-quality semantic parsers for Spanish.

We argued that learner corpora should be leveraged to study overgeneration in grammars systematically, and presented an example where the grammar and the treebank force us to look at the current SRG analysis of Spanish adjectives in the context of the internal structure of the modificand. 

The main avenue for future work apart from general grammar development towards higher coverage and accuracy and lower overgeneration is improving parsing speed. Slow parsing remains a serious problem which requires applying new methods. Recent experiments with training supertaggers for the English Resource Grammar are promising with the speed-up factor of 3 \citep{zamaraeva2023revisiting}, however, training such a supertagger for the SRG will require that the treebanks grow first. That means continuing the research line presented in this paper.

\section{Limitations}
The main limitation of this work is the time cost of grammar engineering and treebanking. Due to the time costs involved, what we present here is work in progress, in the sense that the grammar does not yet cover some syntactic phenomena and some of its existing analyses can be improved: the overgeneration and the ambiguity should be reduced, for example. The results we present are only for sentences up to length 10, and some sentences cannot currently be parsed due to the parser limitations.

\section{Acknowledgments}
We acknowledge the European Union's Horizon Europe Framework Programme which funded this research under the Marie Skłodowska-Curie postdoctoral fellowship grant HORIZON-MSCA‐2021‐PF‐01 (GAUSS, grant agreement No 101063104); the European Research Council (ERC), which has funded this research under the Horizon Europe research and innovation programme (SALSA, grant agreement No 101100615); Grant SCANNER-UDC (PID2020-113230RB-C21) funded by MICIU/AEI/10.13039/501100011033; Xunta de Galicia (ED431C 2020/11); and Centro de Investigación de Galicia ‘‘CITIC’’, funded by the Xunta de Galicia through the collaboration agreement between the Consellería de Cultura, Educación, Formación Profesional e Universidades and the Galician universities for the reinforcement of the research centres of the Galician University System (CIGUS). We also acknowledge grant GAP (PID2022-139308OA-I00) funded by MCIN/AEI/10.13039/501100011033/ and by ERDF ``A way of making Europe''.

\section{References}\label{sec:reference}

\bibliographystyle{lrec-coling2024-natbib}
\bibliographystylelanguageresource{lrec-coling2024-natbib}
\bibliographylanguageresource{languageresource}
\bibliography{srg}


\end{document}